\documentclass[DIV=calc, paper=a4, fontsize=11pt, twocolumn]{article}
\usepackage[utf8]{inputenc}
\usepackage{color}
\usepackage{fancyhdr}
\usepackage{fancybox}
\usepackage{url}
\usepackage{amsmath}
\usepackage{mathtools}
\usepackage[svgnames]{xcolor} 

\usepackage{epigraph}
\usepackage{graphicx, float}
\usepackage{pst-node}
\usepackage[]{algorithm2e}
\usepackage{empheq}
\newcommand{\argmin}[1]{\underset{#1}{\operatorname{arg}\!\operatorname{min}}\;}
\newtheorem{mydef}{Definition}

\usepackage{titling} 

\newcommand{\HorRule}{ \rule{\linewidth}{1pt}} 

\pretitle{\vspace{0pt} \begin{flushleft} \HorRule \fontsize{30}{30} \usefont{OT1}{phv}{b}{n} \selectfont} 

\title{Occurrence Statistics of Entities, Relations and Types on the Web} 

\posttitle{\par\end{flushleft}\vskip 0.5em} 
  
\preauthor{\begin{flushleft}\large \lineskip 0.5em \usefont{OT1}{phv}{b}{sl} \color{Black}} 

\postauthor{\large \usefont{OT1}{phv}{b}{sl} \color{Black} 

\par\end{flushleft}\HorRule} 

\date{April 2014} 

\begin{document}

\begin{titlepage}
    \centering
    \vspace*{\baselineskip}
    \rule{\textwidth}{1.6pt}\vspace*{-\baselineskip}\vspace*{2pt}
    \rule{\textwidth}{0.4pt}\\[\baselineskip]
    {\LARGE \color{DarkBlue} Occurrence Statistics of Entities, Relations and Types on the Web}\\[0.2\baselineskip]
    \rule{ \textwidth}{0.4pt}\vspace*{-\baselineskip}\vspace{3.2pt}
    \rule{\textwidth}{1.6pt}\\[\baselineskip]
    \scshape
    Submitted in partial fulfillment of requirements for the degree of
    Master of Technology \par
    \vspace*{1\baselineskip}
     by \\[\baselineskip]
    {\Large  Aman Madaan \vspace*{1\baselineskip} \\ Under the Guidance of Prof. Sunita Sarawagi\par}
       \begin{figure}[h]
 \centering
 \vspace*{13\baselineskip}
 \includegraphics[bb=0 0 229 220,scale=0.35]{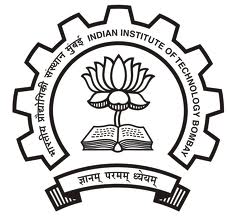}
\end{figure}

    {\vspace*{1\baselineskip}  \itshape Department of Computer Science and Engineering \\ Indian Institute of Technology Bombay\par}
    \vfill
 
    {\scshape April, 2014} \\
    {\large }\par
  \end{titlepage}

\newpage
\maketitle
\begin{abstract}
The problem of collecting reliable estimates of occurrence of entities on the open web forms the premise for this report.
The models learned for tagging entities cannot be expected to perform well when deployed on the web. This is owing to the severe mismatch 
in the distributions of such entities on the web and in the relatively diminutive training data.
In this report, we build up the case for maximum mean discrepancy for estimation of occurrence statistics of entities on the web, taking a review of named entity 
disambiguation techniques and related concepts along the way.
\end{abstract}

\section{Introduction}
\subsection{Problem Statement}
The Internet is a web of mostly unstructured knowledge woven around things. However, these things; people, places, technologies, movies, products, books
etc. are mostly just mentioned by their name, with other crucial bits of information about them scattered around the point of mention. The cosmic scale of 
such unstructured information has stemmed the dream of a semantic web. A web which is aware of the links that make sense, which \emph{understands} what the 
user is looking for and which is gifted with the intelligence of locating the desideratum.
There are several pieces in the puzzle of the semantic web, this report is an attempt to understand one important piece; entities on the web and
their co occurrence statistics.

Given a knowledge base such as Yago or Freebase consisting of entities and relations, and the Web, our goal is 
to attach reliable estimates of the frequency of occurrences on the Web of various entities and relations as 
singletons, pairs (ordered and unordered) in a sentence.  The aim is to collect statistics so as to be able to assign prior probabilities to the set of 
entities and relations that can co-exist in a sentence or a paragraph. These statistics have applications in 
query interpretation and language understanding tasks.  
We can view it as being analogous to statistics in relational catalogs.

\subsection{Named Entity Recognition and Disambiguation}
For collecting the statistics about entities on the web, we need a method to determine which words in the 
free flowing interminable text are of interest, i.e. represent entities. 

Consider the following sentence : 

 \begin{center}
\textcolor{blue}{Michael Jordan is a Professor at Berkeley}
   \end{center}

 We first want to identify all the \textbf{named entities} in the text. The task is called named entity recognition and is 
 formally defined as : 
 \begin{mydef}[Named entity recognition\footnote {from \ref{thewiki}}]
 \label{nerdef}
   Named-entity recognition (NER) (also known as entity identification and entity extraction) is a subtask of information extraction that seeks to locate and classify 
   atomic elements in text into predefined categories such as the names of persons, organizations, locations, expressions of times, quantities, monetary values, percentages, etc.
  \end{mydef}
 but we do not stop at that, we want to link each of the named entities thus recognized to a knowledge base\footnote{The knowledge base is a catalog of entities, like Wikipedia. Refer section [\ref{seckb}]}.
 Thus, our problem has a 2 step solution :

 \begin{itemize}  
  \item Step 1 : \textbf{Identify} entities
  \medskip
  
  \textcolor{green}{Michael Jordan\_PERSON} is a professor at \textcolor{green}{Berkeley\_INSTITUTION} \medskip
  \item Step 2 : \textbf{Link} entities to knowledge bases : 
  \medskip
  
  \textcolor{red}{Michael Jordan\_ENTITY} (\url{http://en.wikipedia.org/wiki/Michael_I._Jordan})  is a professor at  
  \textcolor{red}{Berkeley\_ENTITY} (\url{http://en.wikipedia.org/wiki/University_of_California,_Berkeley})
\end{itemize}

The stanford NER library is a popular choice for recognizing named entities. [\ref{stanfordner}]
\subsubsection{Applications}

In simple terms, disambiguating named entities in the unstructured text imparts a structure to the document. 
We need two more data points to further appreciate the power that such a tool provides to us.
The first is the size of the web. As of 31st March 2014, there are atleast 1.8 billon indexed web pages.[\ref{ws}]
The second is the number of wikipedia entities. The wikipedia statistics [\ref{wikistats}] estimate the number of pages to be
around 32 million. Yago, a catalog of entities made from wikipedia has 12, 727, 222 entities.	
Imparting structure to documents at this magnitude has far reaching implications in the information
extraction and is a bridge towards the hitherto dream of a semantic web.  \\

It is highly recommended that the reader pays \url{http://www.google.co.in/insidesearch/features/search/knowledge.html}, 
the google knowledge graph project, a visit.

\subsubsection{Terminology}
The following terms are widely used in the literature on named entity disambiguation and thus in this article.

\begin{itemize}
 \item \textbf{Mention, Spot} \\
 A piece of text which needs to be disambiguated. For example, the sentence ``\textbf{Amazon} has attracted a lot of visitors''.
 \item \textbf{Entity} \\
 A named entity as defined in the definition \ref{nerdef}. 
 \item \textbf{Candidates} \\
 A set of entities which might be the correct disambiguation for a given mention.
 For example, possible candidates for the sentence above are ``Amazon river'' and ``Amazon.com''.
 \item \textbf{Prior} \\
 Probability of a mention linking to a particular entity. For example, the mention ``Amazon`` may be used
 to refer to the website (say) 60\% of the time.
 \item \textbf{Knowledge base} \\
 A catalog of Entities where an entity is as defined above. For example, Wikipedia or yago.

\end{itemize}

\subsection{A Baseline : Label and Collect}
The baseline which presents itself given the above problem is labeling the corpora with the named 
entities and then collecting the markings, keeping track of which entity was seen when along the way. 
As intuitive as it seems, the method is unlikely to perform well in the present scenario, owing to 
the mismatch in the training and test distribution [\ref{mmd}]. 
Our training data, hand labeled corpora, is paltry in comparison with the massive open web, where such 
systems are supposed to be deployed. This is true even for large training datasets like the Wikipedia.

\subsection{Maximum mean discrepancy}

The observation that we don't really want the individual labels is a first step towards a better solution. 
There are 3 reported methods for direct estimation of class ratios [\ref{mmd}]. We are interested in 
using one of them, maximum mean discrepancy (mmd) for solving the problem in hand.

We introduce mmd and propose a formulation for determining class ratios in section \ref{sec:mmd}.
\subsection{Structure}

\textbf{Section \ref{seckb}} gives an overview of what are knowledge bases. This is important 
since the concept of such repositories of structured knowledge is central to the
report. 

\textbf{Section \ref{secned}} begins with an introduction to the problem of named entity disambiguation, the 
terminology and applications, and goes on to cover the techniques for named entity disambiguation
in some detail. We give and overview of the two broad categories of disambiguation techniques, Local and
global disambiguation.

\textbf{Section \ref{secagg}} begins with a discussion on definition of Aggregate statistics and 
some of their applications. Finally, in section \ref{sec:mmd}, we discuss Maximum mean discrepancy and its 
application for estimating the aggregate statistics over entities.

\section{Structured Knowledge Repositories}
 \label{seckb}
\subsection{What are knowledge bases?}
Before the digital age, Encyclopedias, such as the Encyclopedia Britannica were hailed as the repositories containing
all that is known to the mankind. As the computer age dawned, it didn't take long for people to realize that a lot
can be achieved if somehow all this information could be made available in a digital format.
Wordnet [\ref{wordnet}] was perhaps the first such attempt. As the years passed, the research effort in the field of information extraction and creating 
structured knowledge got a huge pat on the back from the explosion of the web. Wikipedia catalyzed the community, which motivated development 
of structured knowledge bases like dbpedia and yago.

We discuss how knowledge bases fit in the context of named entity disambiguation, and give a list of several
important knowledge bases, along with links to each for the interested reader.

\subsection{Knowledge bases and Named Entity Disambiguation}

Many named entity disambiguation algorithms exploit large knowledge bases.
On the other hand, reliable named entity disambiguators will be conducive towards
fabrication of gargantuan knowledge bases from the open web. We thus see 
a chicken and egg situation here. As is often the case in such standoffs, the cycle is
broken with the help of extensive manual effort. In the present case, Wikipedia helps the
situation.

\subsection{Existing Knowledge Bases}
We give a brief overview of some of the popular knowledge bases.
\subsubsection{Wordnet}
\begin{itemize}
 \item Wordnet has a clean, hand crafted type hierarchy. Well documented APIs, such as the nltk toolkit
(\url{http://www.nltk.org/howto/wordnet.html}) are available for using wordnet for a 
plethora of tasks, such as listing all the senses of a word, finding distances between 
2 concepts and the likes.
 \item Introduction to Wordnet \url{http://wordnetcode.princeton.edu/5papers.pdf}
\end{itemize}

\subsubsection{YAGO}
\begin{itemize}
 \item An attempt to create a knowledge base that combines the clean type hierarchy of 
 wordnet with the huge information that Wikipedia provides. \url{http://www.mpi-inf.mpg.de/yago-naga/yago/}
 has link to an online interface. Refer [\ref{yago}] for details. 
\end{itemize}

 \subsubsection{DBpedia}
 \begin{itemize}
  \item DBpedia \url{http://dbpedia.org/About} extracts information from the Wikipedia into RDF and provides 
  an interface that can be used to ask semantic questions. Users can use SPARQL to ask complicated queries 
  with results spanning several pages. Amazon also provides a DBpedia machine image for the users of AWS.
 \end{itemize}

 \subsubsection{Patty}
 \begin{itemize}
  \item  Patty \url{http://www.mpi-inf.mpg.de/yago-naga/patty/} is a repository of relation patterns. The aim is to 
  create ``Wordnet'' for relations. The authors also create a subsumption hierarchy for the 350, 569 pattern synsets.
  Refer [\ref{patty}] for details.  
 \end{itemize}

\subsubsection{Freebase}
 \begin{itemize}
\item Freebase [\ref{freebase}] relies on crowd sourcing for creation of a rich but clean knowledge base.
The development of Freebase follows the same chain as Wikipedia, with users flagging issues, and
cleaning and augmenting information. Freebase also provides access to itself using web APIs.
\end{itemize}

\section{Named Entity Disambiguation Techniques}
\label{secned}
We have already given an introduction to the problem and the applications in the introduction.
The next section discusses the solutions based on local disambiguation, i.e., figuring 
out the correct entity based on just the local evidences. Section 3.2 discusses the intuition
behind having a global strategy for disambiguation, and the optimization problem that
results from such an objective. The final section summarizes a recent work which 
pragmatically selects global and local evidences, to get the best of both worlds.
\subsection{Local Disambiguation of named entities}
\subsubsection{Introduction}
In local disambiguation, we collect just local evidences for each 
mention for its disambiguation. This was state of the art until the CSAW[\ref{thepaper}]
paper came along. We start by defining the problem and discussing the general form of solutions.
We then provide a short summary of approach followed in Wikify [\ref{wikify}] and the famous Milne and Witten paper [\ref{mw}]. A
solution based on machine learning[\ref{thepaper}] concludes the subsection.

\subsubsection{Problem definition}
We need to disambiguate a mention by collecting the local evidences. 
The evidences can be anything, POS tags, gender information, dictionary lookup 
etc. By local disambiguation, we mean that \textbf{we cannot use the disambiguation
information for any other entities for solving the problem.} 

\subsubsection{Solutions}

Every local disambiguation techniques fall into one of the following two categories[\ref{wikify}]

\begin{itemize}
 \item \textbf{Knowledge based} \\
 Derived from the classical word sense disambiguation literature, this 
 technique depends on the information drawn from the definitions provided by the knowledge base. 
 (See Lesk's algorithm [\ref{lesk}]).
 This is based on the overlap of context with the definitions of each of the candidate 
 senses as given in the knowledge base.
 
 \item \textbf{Machine Learning based} \\
 This method is based on collecting features from the mention and its surroundings, and
 training a classifier to give a verdict on a particular sense being a likely disambiguation
 of a mention. Machine learning based local disambiguation was almost unanimously adopted
 by the ned community as the solution for local disambiguation. AIDA changed the scene 
 by introducing a knowledge based local similarity score which works well.
 
 \end{itemize}

 \subsubsection{Related Work}

\par{Wikify[\ref{wikify}]}
The biggest contribution of this paper is perhaps presenting Wikipedia as the 
catalog against which were supposed to disambiguate. The paper also identifies
two broad methods of doing named entity disambiguation : Knowledge based and 
data based. Since the paper dates back to 2007, when the problem of NED was 
not as established, there are a lot of references to the problem of word disambiguation.

\par{Learning to link with Wikipedia[\ref{mw}]}
This paper defined three different features for disambiguation : 
\begin{itemize}
 \item Commonness  : This is the prior defined in Chapter 1.
 \item Relatedness : Perhaps the biggest contribution of this paper, the relatedness score,
 gives a measure for determining how similar the two entities are. This measure 
 is based on the number of common inlinks to entities in question.
 The relatedness measure as defined here has been used in a lot of works. In fact, all
 the approaches presented in the subsequent subsections use this relatedness score, popular
 as the Milne-Witten score for finding out entity entity similarity.
 This score is defined as follows \\ \\
 $ r(\gamma, \gamma') = \frac{log|g(\gamma) \bigcap g(\gamma')| - log(max\{|g(\gamma)|, |(\gamma')|\})} {log c - log(min\{|g(\gamma)|, |(\gamma')|\})}$ 
 
 Where 
   \begin{itemize}
    \item $g(\gamma)$ : Set of wikipedia pages that link to $\gamma$
    \item $c :$ Total number of Wikipedia pages
    \item $r(\gamma, \gamma') :$ Relatedness of topics $\gamma$ and $\gamma'$
   \end{itemize}\bigskip

\end{itemize}
The algorithm selects a few unambiguous links in the document, and uses the similarity of the candidates
with these unambiguous links as a criteria for disambiguation.
Thus, in some sense, although the technique is not totally local, it shies away from doing anything to maintain
coherence among the entities that are unveiled and thus we do not call this method a ``Global method'', which 
are discussed in the following subsection.

\subsubsection{Machine learning based local disambiguation}
As mentioned, there are primarily two approaches for local disambiguation. 
This subsection discusses a machine learning based local disambiguation method in some detail. This subsection
is based on the local disambiguation approach taken in [\ref{thepaper}].
\par{Definitions}
We first repeat the definitions for quick reference : 
\begin{itemize}
  \item $s$ : Spot, an Entity to be disambiguated (Christian leader John Paul) \bigskip 
  \item $\gamma$ : An entity label value (\url{http://en.wikipedia.org/wiki/Po-pe_John_Paul_II})  \bigskip 
 \item $f_s(\gamma)$ : A feature function that creates a vector of features given a spot and a candidate entity label.
 \end{itemize}
 
 \par{Local compatibility : Feature design} 
 The feature function takes the spot and the candidate as arguments. 
 
\begin{itemize} 
 
 \item The following information about a candidate $\gamma$ is used
\begin{itemize} 
 \item Text from the first descriptive paragraph of $\gamma$
  \item Text from the whole page for $\gamma$
  \item Anchor text within Wikipedia for $\gamma$.
  \item Anchor text and 5 tokens around $\gamma$ 
 \end{itemize}
 
 \item We now have 4 pieces of information about $\gamma$. We take each of these, and apply the following operations with 
 one argument as the spot
    \begin{itemize}
      \item{Dot-product between word count vectors}
      \item{Cosine similarity in TFIDF vector space}
      \item{Jaccard similarity between word sets}
  \end{itemize} 
  \end{itemize} 
 
 Thus, for a candidate - mention pair, we get a total of 12 Features (3 operations, 4 argument pairs). 
 
 In addition to these, we also use a sense probability prior as defined in the introduction. A popular way of 
 obtaining the prior is counting the number of times the spot has been linked to a particular entity. For example, 
 the hypertext ``Linux'' might be linked to the page for the Linux kernel 70\% of the times, and to the page
 for Linux based operating systems rest of the times.

\par{Compatibility Score}
Once we have the features, we train the classifier by using the following optimization objective : 
\begin{itemize}
 \item Local compatibility score between a spot $s$ and a candidate is given by $w^{T}f_s(\gamma)$
 \item $w$ is trained using an SVM like training objective
 \begin{center} $w^{T}f_s(\gamma) - w^{T}f_s(\gamma) \geq 1 - \epsilon_s$ \end{center}
 \end{itemize}
 
 \par{Finding the best candidate}

 \begin{algorithm*}
 \KwData{A Document d}
 \KwResult{Annotated document d' with every mention linked to the best candidate entity}
 \ForEach{mention $m$ in the document} {
  calculate $argmax_{c_m \in \Gamma}w^{T}f_m(c_m)$
  where $\Gamma$ = $c_m$ : $c_m$ is a possible disambiguation of $m$
 }\caption{Local disambiguation}
\end{algorithm*}

 Note that a multi class classifier is not learned for several reasons, all of which can be mapped to 
 the large number of classes.

\subsection{Collective Disambiguation of Named Entities}
\subsubsection{The key intuition}
We have seen several different ``local'' solutions, attempting to solve the problem by collecting evidence
around a mention and then using it to disambiguate. Milne and Witten [\ref{mw}] 
came close to inculcating some sort of coherence, but they couldn't totally build up the intuition. It was after a wait of 2 years that CSAW [\ref{thepaper}] took the game 
to a whole new level by working on the following key intuition :	
\begin{itemize}
  \item A document is usually about one topic \bigskip
  \item Disambiguating each entity using the local clues misses out on a major piece of information : Topic of a page \bigskip
  \item A page is usually has one topic, you can expect all the entities to be \emph{related} to the topic \emph{somehow} \bigskip
  \end{itemize}
  \textcolor{green}{Michael Jackson} : 30 Disambiguations 
  
 \textcolor{green}{John Paul} : 10 disambiguations

  But if they are mentioned on the \textbf{same page}, the page is most likely about Christianity,
  A big hint towards disambiguating \textbf{both} of them.
  
Since the CSAW[\ref{thepaper}] paper, every work on named entity disambiguation includes a 
notion of \emph{Topical coherence} in the solution. 

\subsubsection{Challenges}
Though the notion of topical coherence is very natural and intuitive, there are 
a lot of challenges involved when it comes to actually mapping these intuitions to an optimization
problem.
We present the challenges involved and the solution given by the CSAW team.
 \begin{itemize}
  \item Capturing local compatibility
  \begin{itemize}
   \item \textcolor{blue}{Create a scoring function to rank possible candidates}
  \end{itemize}

  \item Inculcating topical coherence in the overall objective

  \begin{itemize}
   \item \textcolor{blue}{Define Topical coherence}
  \end{itemize}

  \end{itemize}

 Out of these two challenges, various solutions to the problem of capturing the local compatibility are presented in Chapter 2. 
 In this subsection, we focus on the problem of collective disambiguation. 
 
 \subsubsection{The Dominant Topic Model}
  \begin{itemize}
   \item Need to define a collective score based on pairwise topical coherence of all $\gamma_s$ used for labeling. \medskip
   \item The pairwise topical coherence, $r(\gamma_s, \gamma_s')$ is as defined above.\medskip
   \item For a page, overall topical coherence : \begin{center}\medskip
                                                  $\Sigma_{s \neq s' \in S_0}r(\gamma_s, \gamma_s')$
                                                 \end{center}
   \item Can be written as clique potential as in case of node potential\medskip
      \begin{center}
	$exp(\Sigma_{s \neq s' \in S_0}r(\gamma_s, \gamma_s'))$
      \end{center}

  \end{itemize}

  \subsubsection{The Optimization objective}
  With different notations as above, we would like to maximize the following to get the 
  best results.
 \begin{center}
 $\frac{1}{\binom{|S_0|}{2}}\Sigma_{s \neq s' \in S_0}r(\gamma_s, \gamma_s') + \frac{1}{|S_0|}\Sigma_{s \in S_0}w^{T}f_s(\gamma)$
  \\
  \includegraphics[height = 5 cm, scale = 0.3]{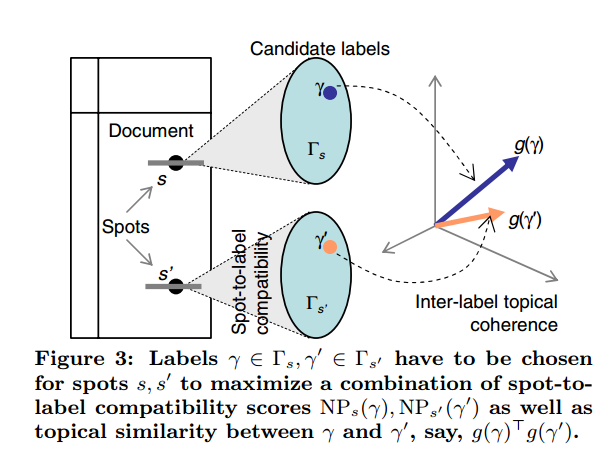}\footnote{Reproduced from [\ref{thepaper}]}
  \end{center}
  In verbose, we want that the entity-entity coherence be maximized, while choosing the disambiguation which is the best.

  \subsubsection{Solving the optimization objective}
  The authors compare 2 different approaches for solving the optimization objective.
  \begin{itemize}
   \item LP rounding approach\bigskip
   
    $|\Gamma|$ + $|\Gamma|^2$ binary variables were introduced. The first set of binary variables decide the candidate that
   each mention takes, and the second set has one binary variable for each possible candidate pair. 
   The authors relax this integer programming to a linear programming and then used rounding with a threshold of 0.5 to
   obtain the best solution.
   
   \item Hill climbing
   
   Starting from all assignments set to NA, assignments are done based on local potentials only. The following figure (
   from the paper) illustrates the process.
   \begin{center}
    \includegraphics[height = 3cm, scale = 0.25]{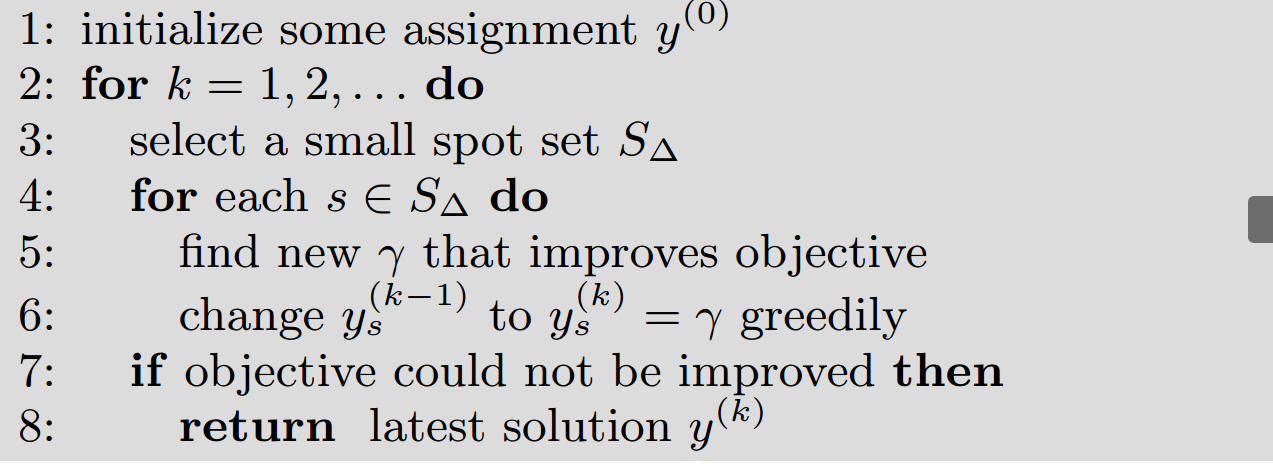}
   \end{center}

  \end{itemize}

\subsection{Pragmatic combination of Local and Global Disambiguations}

\subsubsection{Introduction}

Recall that Chapter 2 was about local disambiguation. In subsection 3, we saw how
global disambiguation can be combined with the overall objective. A recent work, 
Robust disambiguation of named entities in text [\ref{aida}], proposes that 
blindly opting for global disambiguation may not be always right. 
Consider the sentence : ``Manchester will play Madrid in Barcelona''.

All the 3 named entities in the sentence are Cities as well as football clubs.
Collective disambiguation may \emph{coerce} all the three mentions to be 
either football clubs or cities. The work aims to solve this problem by being selective about when to go for collective disambiguation.

\subsubsection{Approach}
This approach first creates a mention to candidate graph. The sample graph for the sentence ``They performed Kashmir 
written by Page and Plant. Page played unusual chord on his Gibson.'' is as shown below : 
 \begin{figure}[H]
 \centering
 \includegraphics[scale=0.23]{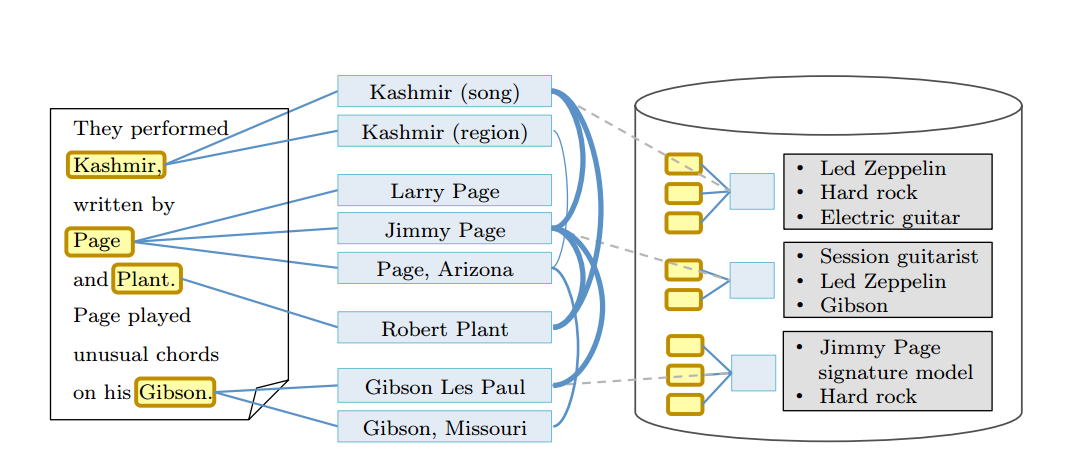}
 \caption{Mention Entity Graph}
\end{figure}

Having created the graph, we need to assign the edge weights. Clearly, there are 2 kinds of edges involved : 
\begin{itemize}
 \item Mention - Entity edge : The authors used a knowledge based approach to assign this weight. This is as outlined in subsection 2.
 The details about this score are given in [\ref{kpsim}].
 \item Entity - Entity edge : Milne witten score as defined in subsection 2 is used for this purpose.
\end{itemize}

With the graph ready, the authors pluck the in a greedy manner such that there is only one edge between each 
mention and entity.

\subsection{Further Readings on Named Entity Disambiguation}

For this report, only a small subset of the papers was selected to cover as much ground as possible.
The following list may be valuable to the interested readers.

\begin{itemize}
\item \textbf{Mining evidences for named entity disambiguation}
The authors discuss a modified LDA model for gathering more words that are important to disambiguate an entity. 
Li, Yang, et al. "Mining evidences for named entity disambiguation." Proceedings of the 19th ACM SIGKDD international conference on Knowledge discovery and data mining. ACM, 2013.
 \item \textbf{We have emphasized on Wikipedia as the catalog. The following work presents a general approach} \\
 Sil, Avirup, et al. "Linking named entities to any database." Proceedings of the 2012 Joint Conference on Empirical Methods in Natural Language Processing and Computational Natural Language Learning. Association for Computational Linguistics, 2012.
 \item \textbf{Large scale named entity disambiguation.} \\
 Cucerzan, Silviu. ``Large-Scale Named Entity Disambiguation Based on Wikipedia Data.'' EMNLP-CoNLL. Vol. 7. 2007.
 \item \textbf{One of the initial works on NED} \\
 Bunescu, Razvan C., and Marius Pasca. "Using Encyclopedic Knowledge for Named entity Disambiguation." EACL. Vol. 6. 2006.
 \item \textbf{Quick entity annotations for short text} \\
 Suchanek, Fabian M., Gjergji Kasneci, and Gerhard Weikum. "Yago: a core of semantic knowledge." Proceedings of the 16th international conference on World Wide Web. ACM, 2007.
\end{itemize}

\section{Distributional Statistics of Named entities}
\label{secagg}
Once you have a catalog of things, it makes sense to ask which of these ``things'' are more important than the others.
In fact, one might extend the question and ask, ``Which pairs (or triples) of these things appear together on the open web?''.
We define several different statistics one might be interested in over these entity catalogs, discuss some applications, 
propose a baseline method and finally, prepare the ground for the next section by giving an outline of a solution 
which is aimed at directly providing us with the statistics we are looking for. 

\subsection{What Statistics?}

\subsubsection{Which sense dominates for an entity?}
For starters, we might want to calculate the number of times a particular ``sense'' of an entity\footnote{Please note that we 
refer to entity in general terms. For example, any object having a YAGO id is an entity} is used. 
For example, the entity Michael Jordan has several disambiguations,  : The Professor, Basketballer and the botinist. 
We want to find out the distribution of occurrences of these senses. We call this number the sense prior.

It is important to note that Entity Prior is different from mention prior, which is the fraction of times a mention 
links to a particular entity. For example, the text ``Gingerbread'' might refer to several different concepts; from perhaps the most famous Android 2.3 to the novel.
Mention prior is to find out how many \emph{Gingerbreads} mentions on the web refer to Gingerbread the Operating system.
Entity sense prior would tell us how frequent is Gingerbread the OS compared with Gingerbread the novel.

\begin{equation}
\tag{1}
\text{Sense Prior}(S_i, E) =  P(E\text{ appears as the $i^{th}$ sense}) = P(S_i | ``E'')
\end{equation}

Where $S_i$ is the $i^{th}$ sense\footnote{$i^{th}$ disambiguation in Wikipedia parlance} of the entity E. 

\subsubsection{How often do the 2 entities appear together?}
A second interesting statistic would be to count how many times do two given entities, taking two given senses appear together.
For example, We might want to know how many times does Nokia \url{http://en.wikipedia.org/wiki/Nokia} appears with Gingerbread \url{http://en.wikipedia.org/wiki/Gingerbread_(operating_system)}

We call these counts Entity bi grams. We note that in contrast to word bi-grams and relational grams [\ref{relgram}], entity bi grams
are symmetric, and there is no obvious use case where we might need to know the order dependent occurrence count of the entities. 
However, such a formulation will lead to a sparse distribution, since each count will have to be normalized by the total number of 
entity bigrams. We thus define the entity bi gram count as follows : 
\begin{equation}
 \tag{2}
 \text{Entity Bi Gram}(E2 | E1) \\  = P(E2\text{ follows }E1) \\ 
	= P(E2 | E1) 
\end{equation}

We propose an application of Entity bi grams for finding out important entities motivated by [\ref{relgram}].

\subsection{Applications}
We list a few applications of the sense prior and outline an application of the entity bigrams.

 \subsubsection{Sense Prior}
 A prior over the sense will be helpful in many applications related to information retrieval. 
 \begin{itemize}
  \item Entity Querying
  \item Knowledge graph based searching
 \end{itemize}

 \subsubsection{Entity Bigrams}
 
 Given an entity, we want to find out other important entities that are related to it.
 For example, given an entity \textbf{Barack Obama, President of the USA}, we need to provide top 10 entities that are
 ``close'' to Barack Obama the President. Since the solution is only a slight modification of the solution 
 presented in [\ref{cohschemas}] for finding out important relations, we only sketch an outline here. 
 
For the entity we are interested in, Say X, create a node. Now attach to the node X all the entities E for which
$P(E|X) >  \epsilon$ where $\epsilon$ is some threshold. Let the weight of the edge be defined as

\begin{equation}
\tag{3}
 P(E|X) + P(X|E) 
\end{equation}

We then apply personalized page rank on the X sub graph, starting with X having a 
page rank of 1 and other nodes having a page rank of 0. We can then sort the nodes
based on the their page ranks upon convergence.

\subsection{Baseline Approach : Label and Collect}
How do we collect the aforementioned statistics?
This question shouldn't be too difficult to answer now. The whole of part 3
was dedicated towards tagging entity mentions in the text. We can use any of the 
methods (for example, AIDA can be set up as a rest service) to tag the corpus, and then iterate over the corpus to collect these statistics
in single pass. 
\subsection{Solution based on estimating class ratios}
While estimating class ratios by doing per mention disambiguation seems pretty intuitive, we are doing more than what we need to do.
We are not interested in what each mention disambiguates to, a count of how many times does a particular entity appears
is the desideratum. There are 3 different methods in the open domain for directly estimating the class ratio[\ref{mmd}] 
, without going through the label and collect route. In particular, [\ref{mmd}] discuss a solution based on maximum mean 
discrepancy and proves some upper bounds on errors. 

If mmd really works, we should expect better estimation of the sense prior and the entity grams. The next section outlines the mmd based 
solution and how mmd may be used to estimate the sense priors for different entities. 

\section{MMD for estimating ratios of named entities in text}
\label{sec:mmd}
This section discusses the MMD approach for direct estimation of class ratios[\ref{mmd}]. We first provide an intuition for the 
solution, follow it up with some results 
\subsection{Introduction}
The following hypothetical example is aimed to capture the gist of class ratio estimation using mmd.
Suppose that in a factory producing balls, there are 3 different ball production machines, (say) A, B and C.
Since neither of the machines is perfect, they do not produce spherical balls. Rather, the balls are
ellipsoids. Thus, for each ball, we have 3 different features corresponding to the three semi-axes. 
Since all the machines are different, they have their own unique view of how balls should look like, 
and thus we expect that the semi axes are a good way of telling the machine which produced a given ball.

Also assume that for all the 3 machines, we also have the most likely (expected) semi axes measures of the balls produced by them.
Let us call these $\phi_a(x), \phi_b(x),$ and $\phi_c(x)$. These are the \emph{expected feature weights}.

Suppose we are given a 150 balls produced from these three machines. For 120 balls out of them, we know the machine 
from which the ball was produced. For the remaining 30 balls, we are asked to give an estimate of how many balls came 
from machine A, B and C. 

How do we do this? Of course, we can learn a classifier from the 120 known instances and then learn the label each of 
the 30 balls and collect counts (label and collect approach). MMD takes the following route to reach the solution.

Suppose we are magically given the true class ratios, say, $\theta_a$, $\theta_b$ and $\theta_c$. Let  
$\phi$ be the average of the semi axes of the 30 balls. Let $\phi'$ be defined as 
\begin{equation}
 \phi' = \phi_a * \theta_a + \phi_b * \theta_b + \phi_c * \theta_c 
\end{equation}

Clearly, we would expect $\phi$ to match $\phi'$. 

Note that we don't really know the $\theta s$, but all is not lost since we know what to look for; 
we look for the thetas that minimize : 
\begin{equation}
 ||\phi_a * \theta_a + \phi_b * \theta_b + \phi_c * \theta_c  - \phi'||^2
\end{equation}

While ensuring that  :
\begin{itemize}
 \item All the $\theta s$ sum to 1.
 \item All the $\theta s$ are non negative.
\end{itemize}

This is the motivation behind MMD for class ratio estimation.
\subsection{MMD Formulation}

With the above example by our side
\subsubsection{Problem Definition}
We reproduce the problem statement from [\ref{mmd}]
 \begin{itemize}
  \item  Let $X = {x \in R_d }$ be the set of all instances and 
  $Y = {0, 1, . . . , c}$ be the  set of all labels.
 \medskip
\item Given a labeled dataset $D(\subset X\text{ x } Y)$, design an estimator that for any given set
$U (\subset X )$ can estimate the class ratios $\theta = [\theta_0 , \theta_1 , . . . , \theta_c ]$
Where  $\theta_y$ denotes the fraction of instances with class label y in U 
 \end{itemize}

\subsubsection{Objective}
\begin{itemize}
  \item Match two distributions based on the mean of features in the hilbert space induced by a kernel K. \medskip
  \item Assume that distribution of features is same in both training and test data
    $P_U (x|y) = P_D (x|y), \forall y \in Y$ \medskip
  \item Thus, the test distribution must equal $Q(x) = \Sigma_{y} P_D (x|y)\theta_y$  \medskip
 \end{itemize}
 
 \begin{itemize}
  \item Let $\bar{\phi}_y$ and $\bar{\phi}_u$ denote the true means of the feature vectors of the y th class and the
unlabeled data \medskip
  \item Suppose we somehow get the true class ratios ${\theta}$. The true mean of the feature vector of the
  unlabeled data can then be obtained by $\Sigma_y\theta_y\bar{\phi}_y$. \medskip
  \item So ideally, $\Sigma_y\theta_y\bar{\phi}_y = \bar{\phi}_u$ \medskip
 \end{itemize}
 The objective thus is
  \begin{equation}
  \argmin{\theta}\Sigma_y{ \in Y\text{  } }|| \Sigma_y\theta_y\bar{\phi}_y - \bar{\phi}_u || ^ {2}  
  \end{equation}
  \begin{center}
  Such that 
  \begin{itemize}
   \item \begin{center} $\forall y, \theta_y \geq 0$ \end{center}
  \item \begin{center} $\sum_{y = 0}^c \theta_y = 1$ \end{center} 
  \end{itemize}
  \end{center}

Interesting discussion on theoretical bounds on the error in the class ratios thus predicted and
methods for learning Kernel can be found in [\ref{mmd}]

\subsubsection{Estimating entity ratios using MMD}
Given a corpus with mentions identified (using, say [\ref{stanfordner}]), we want
reliable estimates of frequency of each of the entities. In this subsection, we gloss over the solution.
\begin{itemize}
\item \textbf{Features}  \\ Each mention has several candidate disambiguations. This gives one way of 
formulating the features. For each mention, we can have a (sparse) feature vector having non zero scores
for the candidates.
\item \textbf{Training data} \\ Can be obtained by splicing the named entity disambiguation pipeline of 
any of the popular named entity disambiguators. [\ref{aidafeature}] discusses how to achieve this for AIDA,
a popular named entity disambiguator. 
\end{itemize}

\section{Conclusion}
The potential of open web can only be harnessed to its full extent by adding structure to it. The process involves 
creating structured repositories derived from the web that can answer interesting questions pertaining to entities
that exist on the web. 

Many such smart applications that rely on structured web will rely on frequencies of occurrence of the former. 
The report has been a buildup to achieving that.
We started by briefing what knowledge bases are. In the second part, we introduced the problem of disambiguating the
mentions of named entities and presented solutions roughly spanning last 8 years of research in the field. 

In the third part, we elaborated on what is meant by aggregate statistics and presented several applications of the same. 
We presented maximum mean discrepancy approach for class ratio estimation via an example and discussed the problem formulation. We briefly outlined how mmd can be applied for
 estimating occurrence statistics of entities.

State of the art approaches for named entity disambiguation brush the figure of 90\% accuracy. It is thus expected
that the focus of the community will now shift to making the process of disambiguation faster and integrating the 
disambiguators in the search pipeline. It remains to be seen how approaches based on direct estimation of entity occurrence 
ratios perform in comparison with the standard tools, both in terms of speed and accuracy.

\section{Acknowledgement}
This report is a summary of selected readings undertaken while working under the guidance 
of Prof. Sunita Sarawagi on application of mmd for collecting occurrence statistics 
of entities on the web. I would like to thank her for the guidance. 
It was immensely helpful in gaining the understanding 
required for writing this report.

Thanks to Mr. Arun Iyer for all the help with understanding maximum mean discrepancy and its implementation.

Lectures by Prof. Soumen Chakarbarti provided useful insights into the problem of named entity disambiguation.

\end{document}